\documentclass[conference]{IEEEtran}
\IEEEoverridecommandlockouts
\usepackage{cite}
\usepackage{amsmath,amssymb,amsfonts}
\usepackage{algorithmic}
\usepackage{graphicx}
\usepackage{textcomp}
\usepackage{xcolor}
\usepackage{multirow}
\usepackage{subcaption}
\usepackage{placeins}
\usepackage{float}
\usepackage{tikz}
\usetikzlibrary{shapes.geometric, arrows.meta, positioning, fit, backgrounds, calc, decorations.pathreplacing}
\def\BibTeX{{\rm B\kern-.05em{\sc i\kern-.025em b}\kern-.08em
    T\kern-.1667em\lower.7ex\hbox{E}\kern-.125emX}}
\begin{document}

\title{Uncertainty-Guided Attention and Entropy-Weighted Loss for Precise Plant Seedling Segmentation}
\author{
\IEEEauthorblockN{Mohamed Ehab}
\IEEEauthorblockA{\textit{Faculty of Computer Science} \\
\textit{October University for Modern Science \& Arts}\\
Giza, Egypt \\
me338484@gmail.com}
\and
\IEEEauthorblockN{Ali Hamdi}
\IEEEauthorblockA{\textit{Faculty of Computer Science} \\
\textit{October University for Modern Science \& Arts}\\
Giza, Egypt \\
ahamdi@msa.edu.eg}
}
\maketitle

\begin{abstract}
Plant seedling segmentation supports automated phenotyping in precision agriculture \cite{JIANG20204152816}. Standard segmentation models face difficulties due to intricate background images and fine structures in leaves. We introduce UGDA-Net (Uncertainty-Guided Dual Attention Network with Entropy-Weighted Loss and Deep Supervision). Three novel components make up UGDA-Net. The first component is Uncertainty-Guided Dual Attention (UGDA). UGDA uses channel variance to modulate feature maps. The second component is an entropy-weighted hybrid loss function. This loss function focuses on high-uncertainty boundary pixels. The third component employs deep supervision for intermediate encoder layers. We performed a comprehensive systematic ablation study. This study focuses on two widely-used architectures, U-Net and LinkNet. It analyzes five incremental configurations: Baseline, Loss-only, Attention-only, Deep Supervision, and UGDA-Net. We trained UGDA-net using a high-resolution plant seedling image dataset containing 432 images. We demonstrate improved segmentation performance and accuracy. With an increase in Dice coefficient of 9.3\% above baseline. LinkNet's variance is 13.2\% above baseline. Overlays that are qualitative in nature show the reduced false positives at the leaf boundary. Uncertainty heatmaps are consistent with the complex morphology. UGDA-Net aids in the segmentation of delicate structures in plants and provides a high-def solution. The results showed that uncertainty-guided attention and uncertainty-weighted loss are two complementing systems.

\end{abstract}


\section{Introduction}
Accurate phenotyping is essential for the success of precision agriculture. Automated systems apply segmentation techniques to plants \cite{JIANG20204152816}, where semantic segmentation addresses the plant pixels and discriminates the background. This allows for leaf area estimations, the determination of growth rates, the prediction of biomass, and the detection of stress. Seedling plant segmentation is described to have challenges \cite{LUO2024172}: background clutter (soil, containers); lighting and color distribution variation; leaf structures, and leaves of fine and delicate nature and growth stages. Conventional CNNs yield poor quality segmentation and provide oversegmentation of the soil. These challenges pose significant barriers to reliable phenotyping in practical situations. Case studies have used existing architectures to propose U-Net (skip connections and spatial resolution), LinkNet (simplified decoder), and Segment Anything Model (SAM) to depth and segmentation jointly \cite{10.1016/j.engappai.2025.111572}. Segmentation of leaves has been improved by implementing attention mechanisms (CBAM, SE-Net) in U-Net \cite{article2}. Uncertainty estimation is used to define unknowns, but in spite of increasing costs, most Bayesian methods and test-time augmentations do not utilize uncertainty during the training process. Some studies employ entropy to evaluate prediction uncertainty, but with a narrow focus on the boundary pixels; nonetheless, the combination of uncertainty, attention mechanisms, and feature refinement has not been addressed in existing studies, and this remains a significant gap. This leads to the development of the present study: UGDA-Net (Uncertainty-guided Dual Attention Network with Entropy-Weighted Loss and Deep Supervision). Three novel components are designed for UGDA-Net:

\begin{itemize}

\item Uncertainty-Guided Dual Attention - uses an uncertainty proxy, in the form of channel variance, to isolate ambiguous features.

\item Hybrid entropy-weighted loss - driven by predictive entropy, emphasizing boundary uncertainties.

\item Deep supervision is derived from outputs at multiple scales of the encoder, offering supplementary training guidance.

\end{itemize}

An elaborate ablation study is performed for two backbones and five grades employing a public-license dataset with carefully annotated images. The three key original contributions are: (i) UGDA with channel variance integrated in the attention mechanism; (ii) a hybrid entropy-weighted loss function and loss that focuses on the most uncertain boundary pixels; (iii) the most comprehensive ablation study to date, measuring the impact of individual components and combinations of components.

\section{Related Work}
This section outlines research pertaining to our methodology. We explain basic segmentation models. We look into attention mechanisms for feature refinement. We consider uncertainty estimation for deep learning. We describe loss functions for class imbalanced segmentation.

\subsection{Semantic Segmentation Architectures}
Fully convolutional networks allow predictions to be made at the pixel level. For example, Ronneberger et al. describe U-Net, a convolutional neural network designed for biomedical image segmentation  \cite{ronneberger2015unetconvolutionalnetworksbiomedical}. A U-Net is designed to be symmetric. There are skip connections from the encoder to the decoder at same level. This enables the network to retain high-resolution information that is useful for making precise predictions. U-Net is particularly useful when training data is scarce. U-Net is to be the baseline in our experiments. Efficient semantic segmentation is addressed by Chaurasia and Culurciello with LinkNet \cite{8305148} . LinkNet is designed with a similar encoder-decoder structure but with dramatically decreased decoder design. In order to minimize the computational load while still achieving satisfactory performance, they included residual connections from encoder to decoder. LinkNet will be included in the ablation study as one of the backbones.

\subsection{Attention Mechanisms}
Mechanisms such as attention optimize certain aspects/ dimensions/ parts of features and representations. Hu et al. \cite{8578843} suggest SE-Net, which enables channel-wise recalibration by means of global pooling and subsequent scaling. Based on this, Woo et al. \cite{10.1007/978-3-030-01234-2_1} present CBAM which conducts channel attention, and then, applies spatial attention. Recent studies combine attention to diminish certainty to segmentation \cite{yin2024uncertainty, ZHOU2024127988}. The UGDA module proposed herein, to the best of the authors’ knowledge, uses channel variance as a proxy to uncertainty to further extend attention mechanisms.

\subsection{Uncertainty Estimation in Deep Learning}
Deep neural architectures lack a measure of predictive certainty and produce point estimates. Monte Carlo Dropout, proposed by Gal and Ghahramani, requires multiple stochastic forward passes at inference time to approximate Bayesian inference \cite{10.5555/3045390.3045502}. Deterministic estimation of both epistemic and aleatoric uncertainties is performed using Gaussian Mixture Models in feature space by Shojaei et al. \cite{10.1007/978-3-031-91767-7_8}. We conduct uncertainty estimation differently by utilizing feature map channel variance and predictive entropy from the output probabilities. This is achievable by performing a single forward pass.

\subsection{Loss Functions for Segmentation}
Binary cross-entropy loss treats every pixel equally and does not address class imbalance. Milletari, Navab, and Ahmadi propose Dice loss to directly optimize the Dice coefficient for volumetric segmentation \cite{article}. Dice loss handles the foreground-background imbalance and addresses small structure segmentation improvement. Hosseini and Soleymani propose a dilated balanced cross-entropy loss that expand class-imbalance enclosing ground-truth class \cite{article22222}. Huang and Sui implement a contour-weighted cross-entropy loss and combine it with separable Dice loss to give emphasis to pixels on the border \cite{huang2024contourweightedlossclassimbalancedimage}. Our loss function integrates binary cross-entropy with Dice loss and weights pixel contributions by predictive entropy. This kind of weighting gives preference to uncertain boundary regions and adapts the loss based on prediction confidence.

\subsection{Summary of Research Gap}
Current architectures have solid segmentation baselines. Attention modules improve the details of features. Uncertainty techniques predict the confidence of segmentation. Class imbalance is solved with specific loss functions. But prior work has separated these techniques. No prior work combines uncertainty attention with focused loss. UGDA-Net is the first to do this. It combines all three techniques into one framework.

\section{Methodology}
This section describes the dataset, the proposed UGDA-Net architecture, the training protocol, and the evaluation metrics. We detail each component of our uncertainty-aware framework.

\subsection{Dataset and Preprocessing}
\begin{figure}[H]
\centering
\includegraphics[width=\columnwidth]{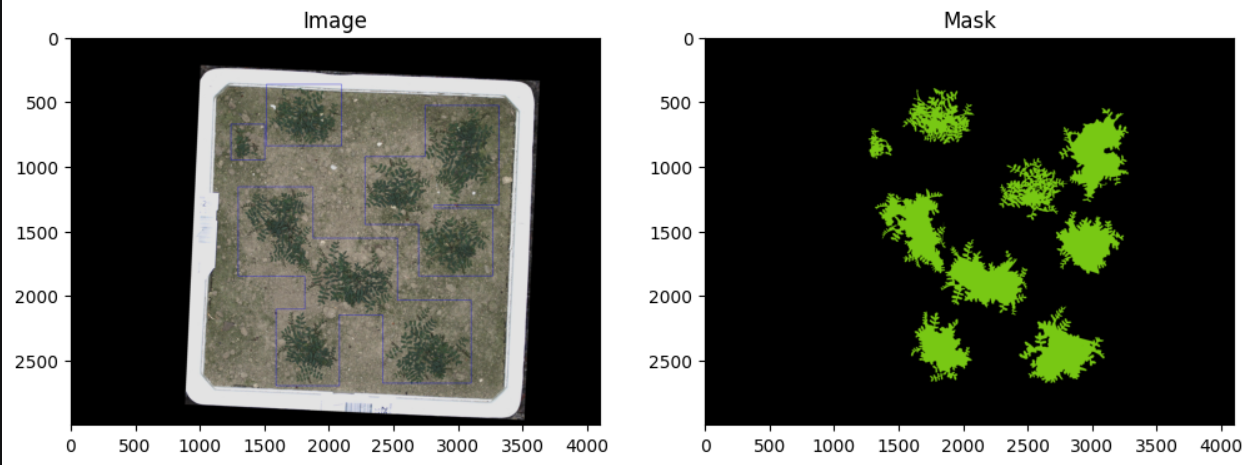}
\caption{Sample from the plant seedling segmentation dataset.}
\label{fig:dataset_sample}
\end{figure}

We utilize a public data set on plant seedling segmentation, which consists of 432 high-resolution RGB images. Each image features a single container containing multiple seedlings. Each image is accompanied by a pixel-level binary mask that indicates plant pixels and background pixels, where background components include soil and container surfaces. Figure~\ref{fig:dataset_sample} shows an example from this dataset.

The dataset was divided into three subsets, with 70\% allocated for training, and 15\% for validation and testing, all with random shuffling. During training, data augmentations are applied, including a resize to 256$\times$256 pixels, and a horizontal flip with a probability of 0.5. Brightness and contrast adjustments and Gaussian blur are applied with probabilities of 0.3 and 0.2, respectively \cite{paper1}. Image channels are normalized with the mean and standard deviation for ImageNet of (0.485, 0.456, 0.406) and (0.229, 0.224, 0.225) respectively. For the validation and test images, only resizing and normalization are applied. Masks are binarized with a threshold of 127: pixels greater than 127 are classified as foreground (class 1) and pixels less than or equal to 127 are classified as background (class 0).

\begin{figure*}[t]
\centering
\resizebox{\textwidth}{!}{%
\begin{tikzpicture}[
    node distance=0.8cm and 1.8cm,
    % Styles
    enc/.style={rectangle, draw=blue!60, fill=blue!5, minimum width=2.0cm, minimum height=1.0cm, align=center, font=\sffamily\small\bfseries, rounded corners=2pt},
    dec/.style={rectangle, draw=blue!60, fill=blue!5, minimum width=2.0cm, minimum height=1.0cm, align=center, font=\sffamily\small\bfseries, rounded corners=2pt},
    ugda/.style={rectangle, draw=orange!80, fill=orange!15, minimum width=2.4cm, minimum height=1.2cm, align=center, font=\sffamily\small\bfseries, text=orange!60!black, rounded corners=2pt},
    ds/.style={rectangle, draw=red!60, fill=red!8, minimum width=2.2cm, minimum height=1.0cm, align=center, font=\sffamily\small, rounded corners=2pt},
    op/.style={font=\sffamily\small, text=black!70},
    arrow/.style={-{Stealth[length=2.5mm]}, thick, black!60},
    skip/.style={-{Stealth[length=2.5mm]}, thick, black!50, dashed},
    dsarrow/.style={-{Stealth[length=2.5mm]}, thick, red!60, dashed},
]

% ==================== INPUT ====================
\node[enc, fill=green!10, draw=green!60] (input) {3};
\node[op, below=0.1cm of input] {Input RGB};

% ==================== ENCODER (Left) ====================
\node[enc, below=0.8cm of input] (enc1) {64};
\node[op, above=-0.1cm of enc1,xshift=40pt] {Conv+BN+ReLU};
\node[op, below=0.1cm of enc1,xshift=40pt] {$H{\times}W$};

\node[enc, below=0.8cm of enc1] (enc2) {128};
\node[op, above=-0.1cm of enc2,xshift=40pt] {Conv+BN+ReLU};
\node[op, below=0.1cm of enc2,xshift=40pt] {$H/2$};
\node[op, left=0.3cm of enc2, rotate=90] {MaxPool};

\node[enc, below=0.8cm of enc2] (enc3) {256};
\node[op, above=-0.1cm of enc3,xshift=40pt] {Conv+BN+ReLU};
\node[op, below=0.1cm of enc3,xshift=40pt] {$H/4$};
\node[op, left=0.3cm of enc3, rotate=90] {MaxPool};

\node[enc, below=0.8cm of enc3] (enc4) {512};
\node[op, above=-0.1cm of enc4,xshift=40pt] {Conv+BN+ReLU};
\node[op, below=0.1cm of enc4,xshift=40pt] {$H/8$};
\node[op, left=0.3cm of enc4, rotate=90] {MaxPool};

\node[enc, below=0.8cm of enc4] (enc5) {512};
\node[op, above=-0.1cm of enc5,xshift=40pt] {Conv+BN+ReLU};
\node[op, below=0.1cm of enc5,xshift=40pt] {$H/16$};
\node[op, left=0.3cm of enc5, rotate=90] {MaxPool};

\draw[arrow] (input) -- (enc1);
\draw[arrow] (enc1) -- (enc2);
\draw[arrow] (enc2) -- (enc3);
\draw[arrow] (enc3) -- (enc4);
\draw[arrow] (enc4) -- (enc5);

% ==================== UGDA MODULES ====================
\node[ugda, right=2.2cm of enc2] (ugda2) {\textbf{UGDA}\\[2pt] \footnotesize Ch(1$\times$1) + Sp(7$\times$7) + $\sigma$(Std)};
\node[ugda, right=2.2cm of enc3] (ugda3) {\textbf{UGDA}\\[2pt] \footnotesize Ch(1$\times$1) + Sp(7$\times$7) + $\sigma$(Std)};
\node[ugda, right=2.2cm of enc4] (ugda4) {\textbf{UGDA}\\[2pt] \footnotesize Ch(1$\times$1) + Sp(7$\times$7) + $\sigma$(Std)};
\node[ugda, right=2.2cm of enc5] (ugda5) {\textbf{UGDA}\\[2pt] \footnotesize Ch(1$\times$1) + Sp(7$\times$7) + $\sigma$(Std)};

\draw[arrow] (enc2.east) -- (ugda2.west);
\draw[arrow] (enc3.east) -- (ugda3.west);
\draw[arrow] (enc4.east) -- (ugda4.west);
\draw[arrow] (enc5.east) -- (ugda5.west);

% ==================== DECODER (Right) ====================
\node[dec, right=2.2cm of ugda5] (dec5) {512};
\node[op, above=-0.1cm of dec5,xshift=40pt] {UpConv};
\node[op, below=0.1cm of dec5,xshift=40pt] {$H/16$};

\node[dec, above=0.8cm of dec5] (dec4) {512};
\node[op, above=-0.1cm of dec4,xshift=40pt] {UpConv + Concat};
\node[op, below=0.1cm of dec4,xshift=40pt] {$H/8$};

\node[dec, above=0.8cm of dec4] (dec3) {256};
\node[op, above=-0.1cm of dec3,xshift=40pt] {UpConv + Concat};
\node[op, below=0.1cm of dec3,xshift=40pt] {$H/4$};

\node[dec, above=0.8cm of dec3] (dec2) {128};
\node[op, above=-0.1cm of dec2,xshift=40pt] {UpConv + Concat};
\node[op, below=0.1cm of dec2,xshift=40pt] {$H/2$};

\node[dec, above=0.8cm of dec2] (dec1) {64};
\node[op, above=-0.1cm of dec1,xshift=40pt] {UpConv + Concat};
\node[op, below=0.1cm of dec1,xshift=40pt] {$H{\times}W$};

\draw[arrow] (ugda5.east) -- (dec5.west);
\draw[arrow] (dec5.north) -- (dec4.south);
\draw[arrow] (dec4.north) -- (dec3.south);
\draw[arrow] (dec3.north) -- (dec2.south);
\draw[arrow] (dec2.north) -- (dec1.south);

% ==================== SKIP CONNECTIONS ====================
\draw[skip] (enc1.east) to[out=0, in=180] node[above, font=\small, pos=0.6] {} (dec1.west);
\draw[skip] (enc2.east) to[out=0, in=180] node[above, font=\small, pos=0.6] {} (dec2.west);
\draw[skip] (enc3.east) to[out=0, in=180] node[above, font=\small, pos=0.6] {} (dec3.west);
\draw[skip] (enc4.east) to[out=0, in=180] node[above, font=\small, pos=0.6] {} (dec4.west);

\node[font=\footnotesize\sffamily, fill=white, inner sep=1pt] at ($(enc1.east)!0.5!(dec1.west) + (3.6,0.4)$) {skip};
\node[font=\footnotesize\sffamily, fill=white, inner sep=1pt] at ($(enc2.east)!0.5!(dec2.west) + (3.6,0.4)$) {skip};
\node[font=\footnotesize\sffamily, fill=white, inner sep=1pt] at ($(enc3.east)!0.5!(dec3.west) + (3.6,0.4)$) {skip};
\node[font=\footnotesize\sffamily, fill=white, inner sep=1pt] at ($(enc4.east)!0.5!(dec4.west) + (3.6,0.4)$) {skip};

% ==================== DEEP SUPERVISION (Side by Side) ====================
% Position aux heads horizontally below encoder columns, spaced apart
\node[ds, below=3.0cm of enc4, xshift=-1.5cm] (ds4) {\textbf{Aux Head 1}\\$w_1{=}0.3$\\[2pt] \tiny Upsample + Loss};
\node[ds, below=3.0cm of enc5, xshift=1.5cm] (ds5) {\textbf{Aux Head 2}\\$w_2{=}0.7$\\[2pt] \tiny Upsample + Loss};

% Draw vertical dashed lines with clear separation
\draw[dsarrow] (enc4.south) -- ++(0,-1.5) -| (ds4.north);
\draw[dsarrow] (enc5.south) -- ++(0,-1.5) -| (ds5.north);

% ==================== OUTPUT ====================
\node[enc, right=1.8cm of dec1, fill=green!10, draw=green!60] (output) {1};
\node[op, above=0.1cm of output] {Conv $1{\times}1$};
\node[op, below=0.1cm of output] {Mask};

\draw[arrow] (dec1.east) -- (output.west);

% ==================== LEGEND (Bottom Right) ====================
\node[draw=black!40, fill=white, rounded corners, font=\sffamily\small, align=left, inner sep=5pt, anchor=south east] 
    at ($(current bounding box.south east) + (-0.5,0.5)$) 
    {%
     \textbf{Legend}\\[2pt]
     \tikz{\node[ugda, minimum width=0.8cm, minimum height=0.4cm] {};} UGDA Module\\
     \tikz{\draw[skip, thick] (0,0) -- (0.5,0);} Skip Connection\\
     \tikz{\draw[dsarrow, thick] (0,0) -- (0.5,0);} Deep Supervision};

\end{tikzpicture}%
}
\caption{UGDA-Net architecture. The blue boxes show feature maps from the encoder/decoder with the number of channels. Orange boxes show where UGDA modules apply uncertainty-guided attention, Stages 2--5. Dashed gray lines indicate skip connections. Red dashed lines show deep supervision head connections with weights $w_1=0.3$ and $w_2=0.7$.}
\label{fig:architecture}
\end{figure*}
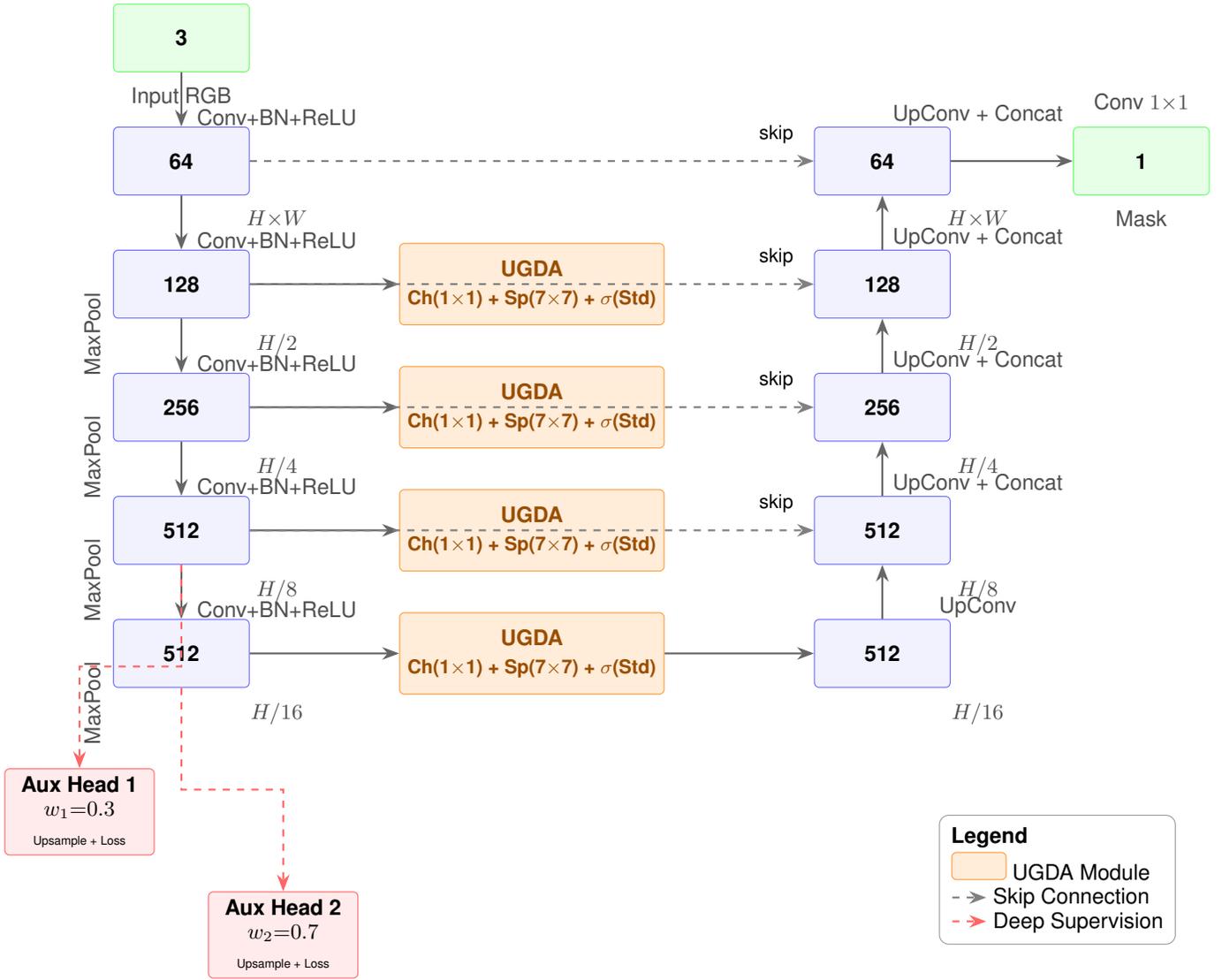

\subsection{Proposed UGDA-Net Architecture}
UGDA-Net (Uncertainty-Guided Dual Attention Network with Entropy-Weighted Loss and Deep Supervision) splits its three novel components into a modified standard encoder-decoder segmentation segmentation network. We assess two possible backbone variants: U-Net and LinkNet. Both use a ResNet34 encoder pre-trained on ImageNet. The complete UGDA-Net framework is shown in Figure ~\ref{fig:architecture}.

\subsubsection{Uncertainty-Guided Dual Attention (UGDA)}
The first novel component is the Uncertainty-Guided Dual Attention module. UGDA features refine maps from the encoder. It integrates channel and spatial attention with an uncertainty signal. Regarding an input feature map, UGDA calculates three attention components. Firstly, channel attention is calculated using global average pooling with two 1×1 convolutional layers. The output from the last layer is filtered through a sigmoid function to generate the channel weights. Secondly, for spatial attention, a 7×7 convolutional layer followed by a sigmoid function is used. Thirdly, an uncertainty map is calculated from the channel-wise standard deviation of the input features. A sigmoid function is used to bound the uncertainty map to [0, 1]. High standard deviation indicates that the features are in disagreement. This means uncertain or ambiguous regions for the features.The attention map after considering all three components is estimated. Channel attention, spatial attention, and the (1 + uncertainty) term are multiplied together, with the uncertainty component also adding additional attention to regions with high feature variance. The module then calculates the refined feature map using a residual connection,

\begin{equation}
Y = X + \gamma \cdot (X \odot A)
\end{equation}

where $\gamma$ is a learnable scalar with an initial value of 0.1. This design choice means that original feature information is preserved, and the attention strength is allowed to be modified by the network during training.We place UGDA modules after every encoder stage, apart from the first one. The shallowest stage keeps its original features.

\subsubsection{Entropy-Weighted Hybrid Loss Function}
The second novel component is an entropy-weighted hybrid loss function. This loss function is a combination of binary cross entropy and Dice loss. It weights pixel contributions based on their predictive entropy.We first calculate pixel-wise entropy from the predicted probabilities. The entropy value quantifies the uncertainty of a prediction. The mid range value of 0.5 has high entropy.  Values close to 0 or 1 have low entropy. High entropy indicates uncertainty around a pixel’s boundary.
We assign a weight map to each pixel as follows:

\begin{equation}
W_i = 1 + \beta \cdot H_i
\end{equation}

We set $\beta = 0.3$ based on empirical validation. This weight map is used to scale the binary cross entropy loss. Uncertain pixels incur a greater penalty. Certain pixels are close to one.The loss from weighted binary cross entropy and Dice loss are combined with coefficients $\lambda_1 = 0.7$ and $\lambda_2 = 0.3$. During training, we implement a warm-up phase. For the first three epochs we use standard binary cross entropy. We then change to the full entropy weighted hybrid loss. This warm-up phase helps to stabilize the early phase of training.

\subsubsection{Deep Supervision}
The third element is deep supervision starting from the intermediate encoder levels. Deep supervision is accompanied by some auxiliary loss. It improves the gradient flow in deep architectures. It also enforces consistency among multi-scale features. We place auxiliary segmentation heads at two of the deepest encoder levels. Each of these heads is one 1×1 convolutional layer that outputs a binary segmentation map. We then upsample these auxiliary outputs to the size of the initial input layer by applying bilinear interpolation. The auxiliary outputs are counted and contribute to the loss. For each auxiliary output we compute the entropy-weighted hybrid loss. We apply a decreasing weight scheme for these auxiliary losses. The deeper layer is assigned 0.7 and the shallower one is assigned 0.3.

The overall training loss for the full UGDA-Net is then

\begin{equation}
\mathcal{L}_{\text{total}} = \mathcal{L}_{\text{main}} + \alpha \sum_{k=1}^{2} w_k \cdot \mathcal{L}_{\text{aux}}^{(k)}
\end{equation}

We have chosen  $\alpha = 0.05$. This value is sufficiently small and balances the contribution from the auxiliary losses without adding too much influence to the main decoder.

\subsection{Training Protocol}
We use PyTorch for every model. Gradient variance is preserved with Glorot uniform initialization \cite{Glorot2010UnderstandingTD}. It also helps with stabilizing the training process. Each model is trained for 40 epochs. We use 16 as the batch size. A learning rate of $1 \times 10^{-4}$ is paird with the AdamW optimizer \cite{Loshchilov2017DecoupledWD}. We use adam \cite{Kingma2014AdamAM}with decoupled weight decay regularization. No learning rate scheduling is used. For lower memory usage and faster training time \cite{micikevicius2018mixedprecisiontraining}, we use automatic mixed precision.We record the model checkpoint with the highes validation Dice coefficient. We use this checkpoint to do the final testing.

\subsection{Evaluation Metrics}
We assess performance in segmentation through two conventional metrics. The two metrics we calculate are the Dice Similarity Coefficient (DSC) and the Intersection over Union (IoU). Each of these two metrics is computed per image, and the average is taken across the test set. We also produce qualitative visualizations. For these visualizations, we present prediction overlays, along with uncertainty heat maps. For overlays, we utilize color codes to indicate the true positives, false positives, and false negatives.

\subsection{Ablation Study Design}
We establish a systematic ablation study. The study determines the impact of each proposed component. We specify five model variants for each backbone architecture:

\begin{itemize}

    \item \textbf{Baseline}: Standard U-Net or LinkNet with standard binary cross-entropy loss.

    \item \textbf{Loss-only}: Baseline architecture with the proposed entropy-weighted hybrid loss.

    \item \textbf{Attention-only}: Baseline architecture with the proposed UGDA module.

    \item \textbf{DS-only}: Baseline architecture with deep supervision auxiliary heads.

    \item \textbf{UGDA-Net (Proposed)}: The complete model combining all three components.

\end{itemize}

We train and evaluate each variant under identical conditions. This design enables fair comparison and attribution of performance gains.

\section{Results}
In this section, we lay out the findings from our ablation study with both the quantitative and qualitative results included. We detail the results for each model variant as seen in the test set. We break down each proposed component and its impact. Finally, we provide some segmentation results and display the uncertainty maps.

\subsection{Quantitative Evaluation}
In Table~\ref{tab:results}, we show the Dice coefficient and the Intersection over Union values for all the variants of the models. Each of the five ablation configurations is tested for both U-Net and LinkNet backbones.

\begin{table}[H]
\centering
\caption{Performance for segmentation on the test set for plant seedling. We report for all model variants the Dice coefficient (DSC) and Intersection over Union (IoU). The best result for each backbone is indicated by bold values.}
\label{tab:results}
\begin{tabular}{llcc}
\hline
\textbf{Backbone} & \textbf{Variant} & \textbf{DSC} & \textbf{IoU} \\
\hline
\multirow{5}{*}{U-Net} & Baseline & 0.4233 & 0.3197 \\
                        & Loss-only & 0.4991 & 0.3743 \\
                        & Attention-only & 0.4323 & 0.3280 \\
                        & DS-only & 0.4144 & 0.3107 \\
                        & \textbf{UGDA-Net (Proposed)} & \textbf{0.5159} & \textbf{0.3905} \\
\hline
\multirow{5}{*}{LinkNet} & Baseline & 0.3519 & 0.2563 \\
                         & Loss-only & 0.4714 & 0.3492 \\
                         & Attention-only & 0.3608 & 0.2563 \\
                         & DS-only & 0.4076 & 0.3048 \\
                         & \textbf{UGDA-Net (Proposed)} & \textbf{0.4840} & \textbf{0.3557} \\
\hline
\end{tabular}
\end{table}

The UGDA-Net proposal establishes the best metrics for both backbones. For U-Net, UGDA-Net yields 9.26 percentage points increase in the Dice coefficient compared to the Baseline. The IoU also increases by an additional 7.08 percentage points. For LinkNet, UGDA-Net improves the Dice coefficient by 13.21 percentage points compared to the Baseline, with an increase of  9.94 percentage points for the IoU. The Loss-only variant shows the greatest improvement for all other variants. For U-Net, Loss-only results in an improvement of Dice by 7.58 percentage points compared to Baseline. For LinkNet, Loss-only results in an improvement of 11.95 percentage points compared to Baseline. The Attention-only variant yields a smaller improvement for U-Net of 0.90 percentage points. For LinkNet, Attention-only yields an improvement of 0.89 percentage points. The DS-only variant results in a small degradation of U-Net. It results in a decrease of Dice by 0.89 percentage points. For LinkNet, DS-only results in an improvement of Dice by 5.57 percentage points. UGDA-Net integrates all three components and therefore outperforms all of the other individual versions. The combination of uncertainty-weighted loss and UGDA attention explains this advancement.

\subsection{Qualitative Analysis}
\begin{figure}[H]
\centering
% Row 1
\begin{subfigure}{0.40\columnwidth}
    \centering
    \includegraphics[width=\linewidth]{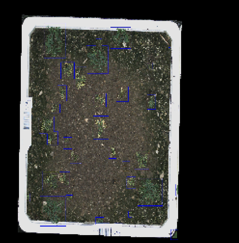}
    \caption{Input image}
    \label{fig:input}
\end{subfigure}
\hfill
\begin{subfigure}{0.40\columnwidth}
    \centering
    \includegraphics[width=\linewidth]{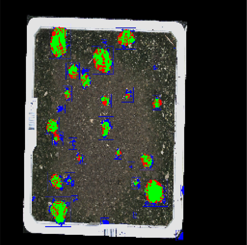}
    \caption{Baseline overlay}
    \label{fig:baseline}
\end{subfigure}

\vskip 0.5em

% Row 2
\begin{subfigure}{0.40\columnwidth}
    \centering
    \includegraphics[width=\linewidth]{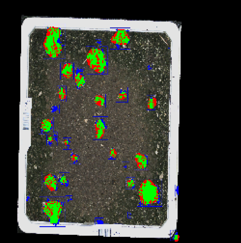}
    \caption{Loss-only overlay}
    \label{fig:loss}
\end{subfigure}
\hfill
\begin{subfigure}{0.40\columnwidth}
    \centering
    \includegraphics[width=\linewidth]{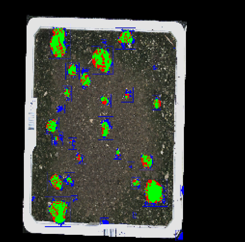}
    \caption{Attention-only overlay}
    \label{fig:att}
\end{subfigure}

\vskip 0.5em

% Row 3
\begin{subfigure}{0.40\columnwidth}
    \centering
    \includegraphics[width=\linewidth]{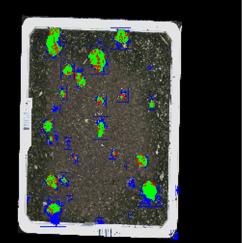}
    \caption{DS-only overlay}
    \label{fig:ds}
\end{subfigure}
\hfill
\begin{subfigure}{0.40\columnwidth}
    \centering
    \includegraphics[width=\linewidth]{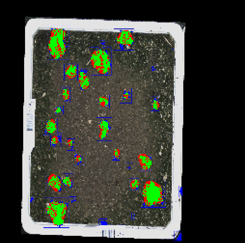}
    \caption{UGDA-Net overlay}
    \label{fig:ugda}
\end{subfigure}

\caption{Results of qualitative segmentation on a sample of a representative test set.
(a) Input RGB image. (b) Baseline overlay. (c) Loss-only overlay.(d) Attention-only overlay. (e) Deep supervision-only overlay. (f) Proposed UGDA-Net overlay. Overlay colors: green = true positive, red = false positive, blue = false negative. UGDA-Net has the best performance in segmentation boundary delineation with the least amount of false positives and false negatives.}
\label{fig:qualitative}
\end{figure}

In Figure~\ref{fig:qualitative}, we have segmented output visualizations for comparison. These segmented outputs are chosen from the test set. For each test example, we present the input image and the ground truth mask, followed by the segmentation predictions from Baseline, Loss-only, and UGDA-Net. The Baseline model produces segmentation masks quite loosely. It overlooks segmentation of fine tips and narrow stems of leaves, and generates false positives in the vicinity of soil regions. The Loss-only variant shows improved following of leaves' boundary. It does segmented capture of leaf structures, while also experiencing a decrease in false positives within background regions. UGDA-Net produces the most valid segmented outputs. It accurately segmented and followed the boundary of fine leaf structures, while also efficiently separating entangled leaves and minimizing false positives on container surfaces. The output images have been generated in different colors. Green pixels denote true positives, while red pixels denote false positives and blue pixels, false negatives. The UGDA-Net output image shows a reduction in red and blue pixels, confirming an increase in segmentation boundary precision. We also present uncertainty heatmaps. These heatmaps display the pixel-wise entropy of the model. Warm color encodes regions with high entropy, and cool color encodes regions with low entropy. The heatmaps associated with UGDA-Net are closely aligned with the leaf boundaries. The model expresses high uncertainty in the complex leaf boundary structures and low uncertainty in background and leaf interiors. This alignment confirms that the entropy-weighted loss is effective.

\subsection{Ablation Insights}
The results of ablation testing indicate some trends. First, the largest contribution is from the entropy-weighted hybrid loss. This validates the use of pixel-wise uncertainty weighting to enhance the segmentation of minute structures. Second, the UGDA module by itself results in very small improvements. But in combination with the uncertainty loss, there is a synergistic collaboration. The attention module is aided with the uncertainty loss. Third, the influence of deep supervision is not the same for the two backbones. U-Net has inherent connections that allow for optimized gradient flow in multiple scales and, therefore, the deep supervision adds redundancy. On the other hand, LinkNet has a less complex decoder and therefore deep supervision is auxiliary in a more meaningful way. The uncertainty-aware training has most improvements in plant seedlings segmentation. However, the full UGDA-Net framework is the combination of all attention, loss weighting, deep supervision.

\section{Discussion}
In the results from the ablation studies, certain trends can be identified. The largest single performance boost came from the entropy-weighted hybrid loss. Validation showed that boundary uncertainty was a factor in segmentation accuracy. The UGDA attention module is required along with the uncertainty-weighted loss, and by itself, yields small improvements. In contrast, the full UGDA-Net is the only configuration that contains both components along with deep supervision, and is therefore the only combination that provides a synergistic effect. U-Net is less benefited from deep supervision than LinkNet, since its skip connections already allow for better gradient flow, and LinkNet also is in greater need of auxiliary supervision. Our deterministic uncertainty estimation is also different from the Bayesian approaches, as it more significantly reduces the forward passes and the associated computations. The present work is not without its limitations. We only use a single dataset and evaluate two different backbones, and we also do not optimize the hyperparameters thoroughly, and we only test with a single size of 256 pixels. We will address the limitations in future work. We will test UGDA-Net on varied plant datasets, along with different transformer backbones, different uncertainty proxies, and we will also optimize it for real-time use.

\section{Conclusion}
This paper introduced UGDA-Net, an original architecture for seedling segmentation that combines Uncertainty-Guided Dual Attention, an entropy-weighted hybrid loss, and deep supervision. We conducted a systematic ablation study on a public dataset of 432 high-resolution images. UGDA-Net improved the Dice coefficient by 9.3\% for U-Net and 13.2\% for LinkNet over baseline models. The entropy-weighted loss provided the largest individual gain, and the UGDA attention module worked synergistically with the loss function. Qualitative results confirmed improved boundary delineation, and uncertainty heatmaps aligned closely with complex leaf edges. UGDA-Net offers a robust framework for uncertainty-aware segmentation and advances automated plant phenotyping in precision agriculture.

\bibliographystyle{plain}
\bibliography{mybibfile}

\end{document}